\documentclass[10pt]{article} 
\usepackage[accepted]{rlc}

\usepackage{amssymb}            
\usepackage{mathtools}          
\usepackage{mathrsfs}           
\mathtoolsset{showonlyrefs}     
\usepackage{graphicx}           
\usepackage{subcaption}         
\usepackage[space]{grffile}     
\usepackage{url}                
\usepackage{enumitem}

\newcommand{\revision}[1]{#1}
\newcommand{\editmark}[1]{{\color{black} #1}}
\newcommand{\editiros}[1]{{\color{black} #1}}
\newcommand{\editiroslsw}[1]{{\color{black} #1}}

\newcommand{\ours}{\textsc{MMN}}
\newcommand{\oursmf}{\textsc{MAH}}

\title{Learning to Navigate in Mazes with Novel Layouts using Abstract Top-down Maps}

\author{Linfeng Zhao\\
    zhao.linf@northeastern.edu\\
    Khoury College of Computer Sciences\\
    Northeastern University
    \And
    Lawson L.S. Wong\\
    lsw@ccs.neu.edu\\
    Khoury College of Computer Sciences\\
    Northeastern University}

\begin{document}

\maketitle

\begin{abstract}
Learning navigation capabilities in different environments has long been one of the major challenges in decision-making.
In this work, we focus on \textit{zero-shot navigation ability} using given abstract $2$-D top-down maps.
Like human navigation by reading a \textit{paper} map, the agent reads the map as an \textit{image} when navigating in a novel layout, after learning to navigate on a set of training maps.
We propose a model-based reinforcement learning approach for this multi-task learning problem, where it jointly learns a \textit{hypermodel} that takes top-down maps as input and predicts the weights of the transition network.
We use the DeepMind Lab environment and customize layouts using generated maps.
Our method can adapt better to novel environments in zero-shot and is more robust to noise.
\end{abstract}

\section{Introduction}
\label{sec:introduction}

If we provide a rough solution of a problem to a robot, can the robot learn to follow the solution effectively?
In this paper, we study this question within the context of maze navigation,
where an agent is situated within a maze whose layout has never been seen before,
and the agent is expected to navigate to a goal without first training on or even exploring this novel maze.
This task may appear impossible without further guidance, but we will provide the agent with additional information:
an abstract $2$-D top-down map, \editiroslsw{treated as an image,} that illustrates the rough layout of the $3$-D environment, as well as indicators of its start and goal locations (``abstract map'' in Figure 1).
This is akin to a tourist attempting to find a landmark in a new city: without any further help, this would be very challenging;
but when equipped with a $2$-D map of environment layout,
the tourist can easily plan a path to reach the goal without needing to explore or train excessively.

In our case, we are most concerned with \textit{zero-shot navigation} in novel environments,
where the agent cannot perform further training or even exploration of the new environment;
all that is needed to accomplish the task is technically provided by the abstract $2$-D map.
This differs from the vast set of approaches based on simultaneous localization and mapping (SLAM) typically used in robot navigation \citep{thrun2005},
where the agent can explore and build an accurate \textit{but specific} occupancy map of \textit{each} environment \textit{prior} to navigation.
Recently, navigation approaches based on deep reinforcement learning (RL) approaches have also emerged,
although they often require extensive training in the same environment \citep{mirowski2016learning, mirowski2018learning}.
Some deep RL approaches are even capable of navigating novel environments with new goals or layouts without further training;
however, these approaches typically learn the strategy of efficiently exploring the new environment to understand the layout and find the goal,
then exploiting that knowledge for the remainder of the episode to repeatedly reach that goal quickly \citep{jaderberg2017}.
In contrast, since the solution is essentially provided to the agent via the abstract $2$-D map,
we require a more stringent version of zero-shot navigation, where it should \emph{not} explore the new environment;
instead, we expect the agent to produce a near-optimal path in its \textit{first} (and only) approach to the goal.

\begin{figure}[!t]
    \centering
    \includegraphics[width=.5\linewidth]{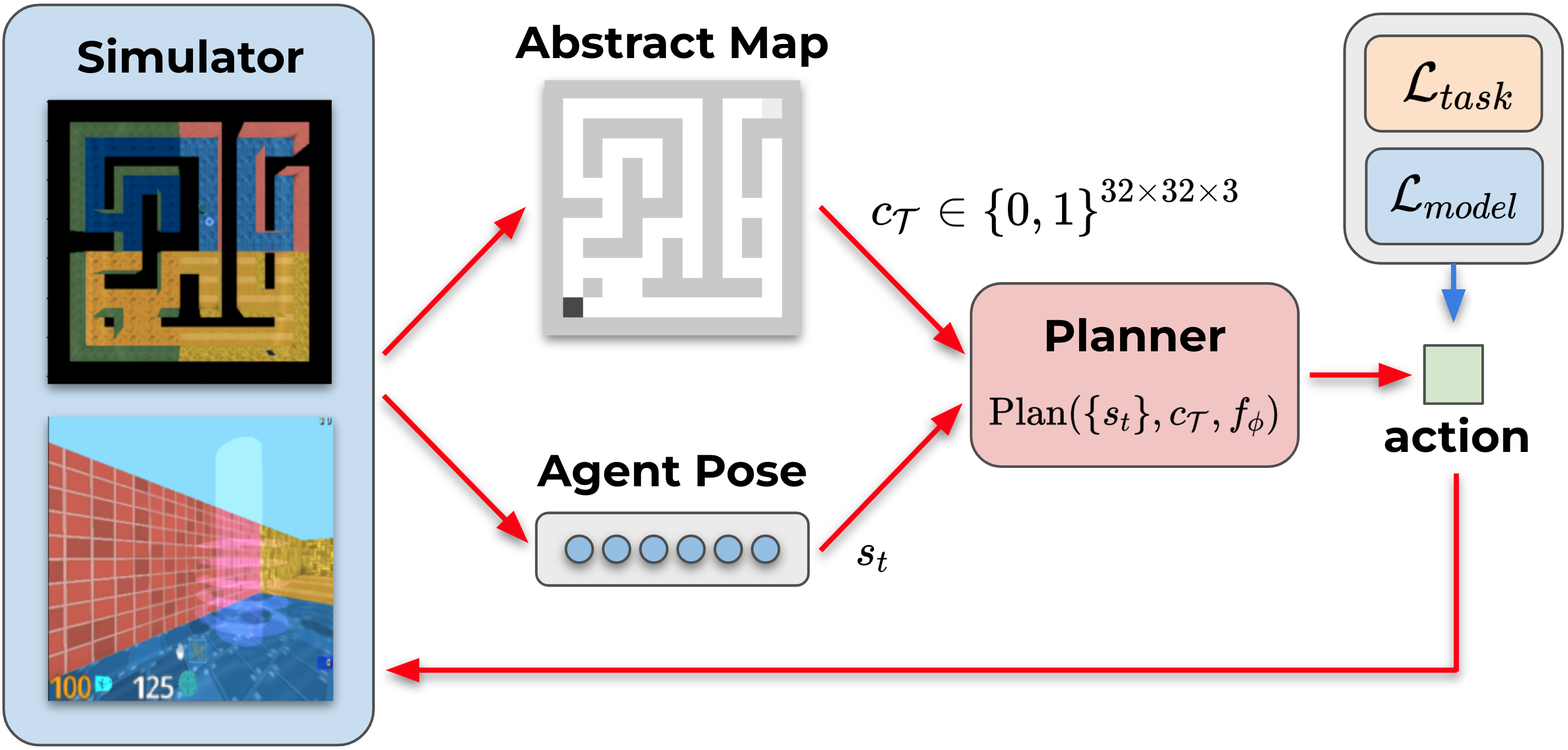}
    \caption{
    \small
    We develop an agent that can perform zero-shot navigation on unseen maps $\mathcal{T}$ (in DeepMind Lab, blue box), without needing to first explore the new $3$-D environment.
    Instead, the agent is given the \textit{top-down view} as additional guidance: an abstract $2$-D occupancy map, and a goal and start position (bottom-left black dot and top-right gray dot).
    The map provides a rough solution solution, the path cannot be directly followed due to the continuous nature of the agent's environment, as well as unknown map scale, inaccuracies in the map, and noisy localization.
    }
    \vspace{-10pt}
    \label{fig:0-overview}
\end{figure}

\editiroslsw{The solution to navigation using the provided abstract map seems obvious: we should localize ourselves on the abstract map (image), plan a path, and simply follow it. However, this approach suffers from a key difficulty: determining the correspondence between $2$-D image maps and $3$-D environments. It is not obvious how to execute the abstract plan in practice because the state and action spaces are completely different, and may even be discrete in the abstract map but continuous in the real environment.}

\editiroslsw{Instead, in this paper we explore an alternative approach that avoids explicitly localizing and planning on the abstract map. The key idea is to plan in a learned model that only considers the abstract map (and start/goal information) as contextual input, but does not directly plan on the map image itself. Specifically, we propose learning a \emph{task-conditioned hypermodel} that uses the abstract map context to produce the \emph{environment-specific parameters} (weights) of a latent-state transition dynamics model. We then perform planning by using sampling-based forward search on this task-specific dynamics model. Importantly, although the learned transition model operates in latent state space, it uses the agent's original action space, so that planned trajectories can be directly executed in the environment, without needing to solve the aforementioned correspondence problem. The hypermodel and the state encoder are learned in an end-to-end fashion, using loss functions that assess whether the learned components were able to support effective planning.}

\editiroslsw{We refer to our method as the Map-conditioned Multi-task Navigator (\textbf{\ours}). We start with a model-based RL algorithm, MuZero \citep{schrittwieser2019mastering}, and introduce the above task-conditioned hypermodel based on HyperNetworks \citep{ha2016hypernetworks}. To tackle challenges in training, we additionally introduce an $n$-step generalization of Hindsight Experience Replay (HER) \citep{andrychowicz2017hindsight} and an auxiliary hypermodel loss.}
\editiroslsw{Additionally, we introduce a model-free RL baseline, named Map-conditioned Ape-X HER DQN (\textbf{\oursmf}). This method builds upon DQN \citep{Mnih2015, horgan2018distributed} and augments the input with the provided abstract map, and uses standard single-step HER.}

In experiments performed in DeepMind Lab \citep{beattie2016deepmind}, a $3$-D maze simulation environment shown in Figure 1,
we show that both approaches achieve effective zero-shot navigation in novel environment layouts, though the model-based \textbf{MMN} is significantly better at long-distance navigation.
\revision{Additionally, whereas a baseline approach using deterministic path planning and reactive navigation quickly fails when the map is inaccurate or localization is noisy, our experiments suggest that \textbf{MMN} is significantly more robust to such noise.}

\section{Related work}
\label{sec:related}

Navigation is widely studied in robotics, vision, RL, and beyond;
to limit the scope, we focus on zero-shot navigation in novel environments, which is most relevant to this work.
This excludes traditional approaches based on SLAM \citep{thrun2005},
since those methods need to explicitly build a map before navigation,
and the map can only be used for the corresponding environment and cannot be transferred to other layouts.
Learning-based methods (e.g., \citet{mirowski2016learning,mirowski2018learning}) also require extensive training data from the same environment;
they demonstrate generalization to new goals in the environment, but not transfer to new layouts.
\citet{jaderberg2017,chen2019learning,Gupta2019,Chaplot2020} demonstrate agents that learn strategies to explore the new environment
and potentially build maps of the environment during exploration;
in contrast, we are interested in agents that do not need to explore the new environment.
\citet{Gupta2019}
learns to exploit semantic cues from its rich visual input, which is orthogonal to our work since we use the state directly.
Other domains such as first-person-shooting games also involve agents navigating in novel environments
\citep{Lample2017,Dosovitskiy2017,Zhong2020}, but since navigation is not the primary task in those domains,
the agents may not need to actually reach the specified goal (if any).
Most closely related to our work is \citet{brunner2018teaching}, who also use $2$-D occupancy maps as additional input and perform experiments in DeepMind Lab.
Their approach is specific to map-based navigation, whereas our methodology aims to be less domain specific.
\citet{huang2021continual} also use HyperNetworks on robot manipulation tasks.

\editiros{
Our work is an instance of \textit{end-to-end model-based planning}~\citep{tamar2016value,oh2017value,schrittwieser2019mastering,zhao_integrating_2022}.
It has also been referred to as \textit{implicit model-based planning} since the model is learned implicitly, compared to approaches with explicit models~\citep{xie2020deep,kumar2024practice}.
It rolls out trajectories using a learnable transition model and jointly trains the \textit{value and policy} networks along with the \textit{transition} network.
This is different from \textit{decoupled} model learning and planning, such as \textit{Dyna-style} \citep{pong2018temporal}.
One important distinction in end-to-end planning is whether the gradients are passed through the \textit{planning computation}.
For example, MuZero \citep{schrittwieser2019mastering} uses sampling-based search method, Monte Carlo tree search (MCTS), that is hard to differentiate though.
Other sampling-based approaches include \citep{hafner2018learning,chua2018deep,zhao2024equivariant,zhao2023euclidean}.
Another thread of work includes Value Iteration Networks and its variants \citep{tamar2016value,lee_gated_2018,zhao_integrating_2022,zhao_scaling_2023,zhao2023e2equivariant,}, which iteratively applies Bellman operators and is easily differentiable.
They have also been used in end-to-end navigation, including CMP \citep{Gupta2019} and DAN \citep{karkus_differentiable_2019}.
However, they are limited to grid-like structure as the VIN backbone is $2$-D convolution.
Additionally, a body of work \citep{parisotto2017neural,banino2018vector,fortunato2019generalization,wayne2018unsupervised,ma2020discriminative,park2022learning,zhao2022toward,howell2023equivariant,Jia2024openvocabulary} studies learning structured latent models or representations useful for planning, such as using symmetry. 

Our method is based on MuZero \citep{schrittwieser2019mastering}, which has only been used on single-map/goal navigation because it learns purely from rewards. We augment the approach with task conditioning (map and goal) to generalize to new layouts. 
\citet{moro2022goal} also introduced goal-relabeling for AlphaZero and applied it in $2$-D navigation; however, 
AlphaZero requires a given model, whereas MuZero jointly learns and plans with a model. 
}

\section{Problem statement}
\label{sec:problem}

We consider a distribution of navigation tasks $\rho(\mathcal{T})$.
Each task is different in two aspects: map layout and goal location.
(1) \textit{Abstract map}. The layout of each navigation task is specified by an abstract map. Specifically, an abstract map $m\in \mathbb{R}^{N \times N}$ is a $2$-D occupancy grid, where cell with $1$s (black) indicate walls and $0$s (white) indicate nagivable spaces. A cell does not correspond to the agent's world, so the agent needs to learn to localize itself on an abstract $2$-D map (i.e., to know which part of map it is currently at). We generate a set of maps and guarantee that any valid positions are reachable, i.e., there is only one connected component in a map.
(2) \textit{Goal position.} Given a map, we can then specify a pair of start and goal position.
Both start and goal are represented as a ``one-hot'' occupancy grid $g\in \mathbb{R}^{2\times N \times N}$ provided to the agent. 
For simplicity, we use $g$ to refer to both start and goal,
and we denote the provided map and start-goal positions $c=(m,g)$ as the \textit{task context}. 

We formulate each navigation task as a goal-reaching \textit{Markov decision process} (MDP), consisting of a tuple $\langle \mathcal{S}, \mathcal{A}, P, R_{\mathcal{G}}, \rho_0 ,\gamma \rangle$, where $\mathcal{S}$ is the state space, $\mathcal{A}$ is the action space, $P$ is the transition probability function $P:\mathcal{S\times A}\rightarrow \Delta(\mathcal{S})$, $\rho_0 = \rho(s_0)$ is the initial state distribution, and $\gamma \in (0,1]$ is the discount factor.
In the learning, we assume transitions are deterministic.
For each task, the objective is to reach a subset of state space $\mathcal{S}_\mathcal{G} \subset \mathcal{S}$ indicated by a reward function $R_{\mathcal{G}} : \mathcal{S} \times \mathcal{A} \rightarrow \mathbb{R}$. We denote a task as $\mathcal{T} = \langle P , R_{\mathcal{G}}, \rho_0 \rangle$, since a map and goal specify the dynamics and reward function of a MDP, respectively.
In the episodic goal-reaching setting, the objective is typically not discounted ($\gamma=1$) and the reward is $-1$ for all non-goal states, i.e., $R_\mathcal{G}(s,a)=- \mathbb{I}[s \neq g]$ for $g\in \mathcal{S}_\mathcal{G}$.

\revision{We emphasize that although the abstract map's occupancy grid corresponds to the environmental layout, the correspondence between abstract ``states'' (grid cells) and agent states (pose and velocity) is not known in advance, and likewise for actions (grid-cell transitions vs. forward/backward/rotate). Furthermore, the learned correspondence may not be reliable due to inaccuracies in the abstract map and localization error.}

\begin{figure}[h]
    \centering
    \vspace{-10pt}
    \includegraphics[width=0.4\linewidth]{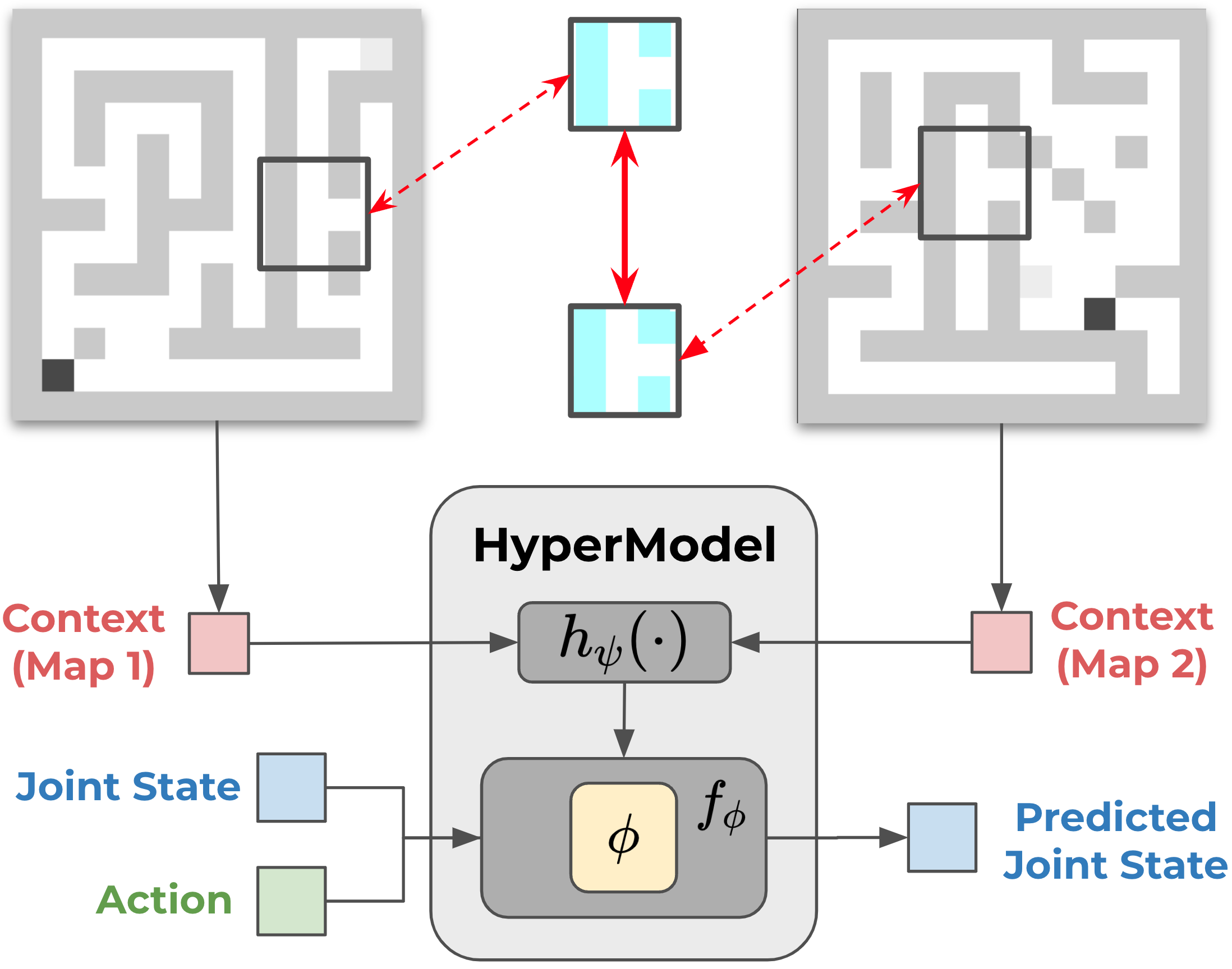}
    \caption{
    \small
    Applying the hypermodel $h_\psi$ on map $m_1$ and $m_2$ outputs two sets of transition network weights $\phi_1=h_\psi(m_1,g_1)$ and $\phi_2=h_\psi(m_2,g_2)$.
    Each transition network uses their weight $\phi_i$ to predict the next state $f(s,a; \phi_i)=s'$, illustrated at the bottom.
    Since the maps may share local patterns at some scales (illustrated by the cropped $3\times3$ patches in light blue), they can be captured by the hypermodel $h_\psi$.
    }
    \vspace{-10pt}
    \label{fig:figure2-hypermodel}
\end{figure}

\section{Learning to navigate using abstract maps}
\label{sec:approach}

\editiros{
This section presents an approach that can effectively use abstract maps (in image form) by end-to-end model-based planning based on MuZero \citep{schrittwieser2019mastering}.
We expect the agent to be able to efficiently \textit{train} on multiple maps as well as \textit{generalize} to new maps.
}

\editiros{
This poses several technical challenges.
}
\textit{(i)} A local change in map may introduce entirely different environment structure, so we need the model and planner to adapt to the task context in a different way than conditioning on state, and not directly condition on the entire task context.
\textit{(ii)} During training, we can only rely on a very small proportion of training tasks (e.g., $20$ of $13\times 13$ maps).
\editiros{This requires compositional generalization from existing map patches to novel combinations of patches.}
\textit{(iii)} The reward signal is sparse, but model learning is done jointly and purely relies on reward signal.
\editiros{
To this end, we first introduce the idea of using a \textit{hypermodel} that learns to predict weights of transition model, instead of state output directly, to tackle \textit{(i)} and \textit{(ii)}.
For challenge \textit{(iii)}, we use the idea from Hindsight Experience Replay (HER) \citep{andrychowicz2017hindsight} to reuse failure experience and also add an auxiliary loss of predicting transitions (described in Appendix~\ref{sec.appendix-approach}).
}

\subsection{Task-conditioned hypermodel}

\editiros{
Our goal is to create a transition model that accurately handles various map inputs, enabling planning in 3D environments with arbitrary layouts. 
}
In a single-task training schema, a straightforward approach would be to learn a parameterized transition function $f_i(s, a)$ for each individual map.
\editiros{
However, we aim to leverage shared knowledge between navigation tasks, where maps often exhibit common local patterns and require the ability to generalize to recombination of known patterns.
}
For instance, in Figure~\ref{fig:figure2-hypermodel}, moving right on the center of the box in the left map shares computation with the right map.
By enabling the agent to recognize these local computational patterns, \editiros{it can transfer to new tasks by compositional generalization.}

We propose to build a \textit{meta} network $h_\psi$, or \textit{hypermodel}, to learn the ``computation'' of the transition model $f_\psi$ simultaneously for all maps with abstract 2-D maps as input.
The transition model for task $\mathcal{T}$ (map-goal pair) is a function $f_i$ that maps current (latent) state and action to a next (latent) state.
We parameterize a transition function $f_i$ as a neural network with its parameter vector $\phi_i$.
The set $\left\{ f_i \right\}$ represents transition functions of all tasks belonging to a navigation schema (e.g., a certain size of map), and these tasks have similar structure.
This set of transition functions/networks are characterized by the context variables $c=(m,g)$, i.e., the abstract $2$-D map and goal.\footnote{Concretely, a task context $c \in \mathbb{R}^{4\times N \times N}$ has four components: downsampled global occupancy map, cropped local occupancy map, and one-hot goal and start occupancy maps; $N$ is downsampled size.} This implies that parameter vectors $\phi_i$ live in a low-dimensional manifold.
Thus, we define a mapping $h: \mathcal{C} \rightarrow \Phi $ that maps the context of a task to the parameter vector $\phi_i$ of its transition function $f_i$, predicting state $s'$ and reward $r$.
We parameterize $h$ also as a network with parameter $\psi$:\footnote{We only predict weights of the transition model $f_\phi:\mathcal{S} \times \mathcal{A} \rightarrow \mathcal{S}$ which operates on a latent state space. The mapping from environment observations to latent states $e: \mathcal{O} \rightarrow \mathcal{S}$ is not predicted by a meta network. Since the latent space is low-dimensional, it is feasible to predict weight matrices of a transition network for it.}
\begin{equation}
    h_\psi: c \mapsto \phi, \quad f_\phi : s, a \mapsto s', r.
\end{equation}
This can be viewed as soft weight sharing between multiple tasks. 
It efficiently maps low-dimensional structure in the MDP, specified by the map, to computation of the transition model.
It may also be viewed as a structured \textit{learned} ``dot-product'' between task context $c_\mathcal{T}$ and state and action $s_t, a_t$ to predict the next state.
The idea of predicting the weights of a main network using another \textit{meta}-network is also known as HyperNetworks~\citep{ha2016hypernetworks,von2019continual}.

\begin{figure}[h]
    \centering
    \includegraphics[width=\linewidth]{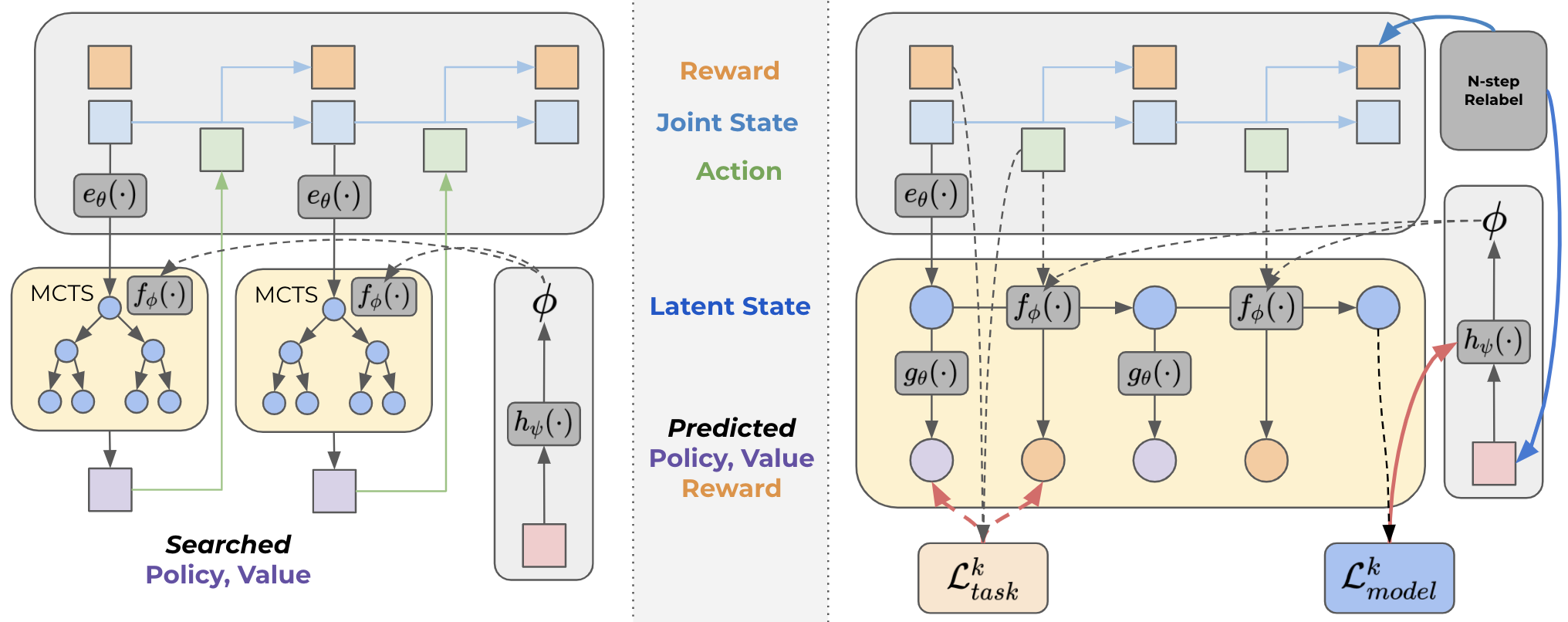}
    
    \caption{
    \small
    The planning/learning process. Yellow boxes indicate predictions; grey boxes come from actual interactions.
    \editiros{\textit{(Left) Inference: search with learned model.}}
    Applying MCTS with hypermodel to search for policy and value, and act with a sampled action.
    \editiros{\textit{(Right) Training: building learning targets.}}
    Computing targets and backpropagating from loss.
    The dark blue line indicates $n$-step relabelling. We only illustrate backpropagation for one reward node for simplicity.
    The solid red line shows the gradient flow from auxiliary model loss to the meta-network's weight $\psi$.
    The dashed red line is the gradient from task loss.
    }
    \vspace{-10pt}
    \label{fig:inference-training}
\end{figure}

\subsection{Planning using a learned hypermodel}

Equipped with a map-conditioned model, we use it to search for actions according to the map layout and goal location: $(a^1, ..., a^k)=\text{Plan}(\{s_i \}, c, f_\phi)$.
\editiros{We follow MuZero \citet{schrittwieser2019mastering}} to use Monte-Carlo tree search (MCTS) to search with the learned hypermodel $f_\phi$.
\editiros{The planner needs to act based on different task inputs, which necessitates a task-dependent value function that differs from the single-task setup in MuZero. 
Consequently, the planner $\text{Plan}({s_i }, c, f_\phi)$ must strongly correlate its computation with the map and goal input $c=(m,g)$, which presents a challenge for model-free reactive agents.}
\editiros{As shown in Figure~\ref{fig:inference-training} (left), we begin by encoding the observed joint state $o_t$ into a latent space $s_t$ using the learned encoder $e_\theta(o_t)$.
This serves as the root node of the search tree. 
To predict the next state given a latent state and a candidate action, we use the hypermodel $f_\phi$. For each state (blue circle nodes), we use another network $g_\theta(s_t, c)$ to predict the policy $\pi_t$ and value function $v_t$ (not shown).
These networks guide the search, where the value network estimates the future value and the policy network provides candidate actions for rollout in MCTS (blue circles), as described in \citet{schrittwieser2019mastering}. During training, they are trained to minimize the loss with searched values and actions.
Once a number of MCTS simulations are completed (yellow rounded boxes), we backup the statistics to the root node and sample an action (green boxes) from the searched action distribution (purple boxes).}
The trajectory and corresponding abstract map and goal $\left(c_\mathcal{T}, \left\{ s_t, a_t, r_t, s_{t+1} \right\}_t \right)$ are saved to a centralized replay buffer for training.

At \textbf{zero-shot evaluation} time, given a new abstract map, we plan with the trained hypermodel: (1) given a map and goal $c_\mathcal{T}=(m_\mathcal{T},g_\mathcal{T})$, at the beginning of the episode, compute the hypermodel weights $\phi=h(c;\psi)$ by applying the meta-network on the task context $c_\mathcal{T}$, (2) start MCTS simulations using the hypermodel $f(s,a; \phi)$ for latent state predictions, (3) get an action and transit to next state, and go to step (2) and repeat.
Moreover, if we assume access to a \textit{landmark oracle} on given maps, we can perform \textbf{hierarchical navigation} by generating a sequence of local subgoals $\{(m, g_i) \}_{i=1}^n$, and plan to sequentially achieve each landmark; see Section~\ref{sec:hierarchical} for more details.

Figure~\ref{fig:inference-training} (right) shows our goal-relabeling scheme and loss functions; see Appendix~\ref{sec.appendix-approach} for details.

\section{Experiments}
\label{sec:experiments}

In the experiments, we assess our method and analyze its performance on DeepMind Lab \citep{beattie2016deepmind} maze navigation environment.
We focus on zero-shot evaluation results.

\subsection{Experimental setup}

We perform experiments on DeepMind Lab \citep{beattie2016deepmind}, an RL environment suite supporting customizing 2-D map layout.
As shown in Figure \ref{fig:0-overview}, we generate a set of abstract $2$-D maps, and use them to generate $3$-D environments in DeepMind Lab.
Each cell on the abstract map corresponds to $100$ units in the agent world.
In each generated map, all valid positions are reachable, i.e., there is only one connected component in the map.
Given a sampled map, we then generate a start-goal position within a given distance range.
Throughout each task, the agent receives the abstract map and start/goal location indicators, the joint state vector $o \in \mathbb{R}^{12}$ (consisting of position $\mathbb{R}^3$, orientation $\mathbb{R}^3$, translational and rotational velocity $\mathbb{R}^6$), and reward signal $r$.
The action space is \{forward, backward, strafe left, strafe right, look left, look right\}, with an action repeat of $10$.
This means that, at maximum forward velocity, the agent can traverse a $100 \times 100$ block in two steps, but typically takes longer because the agent may slow down for rotations.

\paragraph{Training settings}
We train a set of agents on a variety of training settings, which have several key options: 
(1) \textit{Map size}. We mainly train on sets of $13 \times 13, 15 \times 15, 17 \times 17, 19 \times 19, 21 \times 21$ maps.
One cell in the abstract map is equivalent to a $100 \times 100$ block in the agent's world.
(2) \textit{Goal distance}. During training, we generate local start-goal pairs with distance between $1$ and $5$ in the abstract map. 
(3) \textit{Map availability}. For each map size, we train all agents on the same set of $20$ generated maps, with different randomly sampled start-goal pairs in each episode.

\paragraph{Evaluation settings}
We have several settings for evaluation: 
(1) \textit{Zero-shot transfer}. We mainly study this type of generalization, where the agent is presented with $20$ unseen evaluation maps, and has to navigate between randomly generated start-goal pairs of varying distances.
(2) \textit{Goal distance} on abstract map.
We consider both \textit{local} navigation and \textit{hierarchical} navigation. In the \textit{local} case, we evaluate on a range of distances ($[1,15]$) on a set of maps, while in the \textit{hierarchical} case, we generate a set of landmarks with a fixed distance of $5$ between them and provide these to agents sequentially.
(3) \textit{Perturbation}. To understand how errors in the abstract map and in localization affects performance, we evaluate agents with maps and poses perturbed by different strategies.

\paragraph{Evaluation metrics}
We mainly report success rate and (approximate) SPL metric~\citep{anderson2018evaluation} with $95\%$ confidence intervals (higher SPL is better).
We report results from fully trained agents to compare asymptotic performance; no training is performed on evaluation maps.

\paragraph{Methods}

\editiros{
We compare our model-based approach against two model-free baselines. 
\begin{enumerate}[leftmargin=*, nosep]
	\item \textit{Map-conditioned Multi-task Navigator} (\textbf{\ours}), model-based. Our map-conditioned planner based on MuZero and improved with $n$-step HER and multi-task training.
	\item \textit{Map-conditioned Ape-X HER DQN} (\textbf{\oursmf}), model-free. Based on Ape-X DQN \citep{horgan2018distributed} and single-step HER \citep{andrychowicz2017hindsight}, \textit{conditioned} on map and goal.
	\item \textit{Single-task Ape-X HER DQN} (\textbf{DQN}$^\dagger$). 
        Similar to above, but no task context $c$ provided. 
	\item \textit{Random}, a reference of the navigation performance.
\end{enumerate}
}

\begin{figure}[t]
    \centering
    \begin{subfigure}{0.45\linewidth}
		\includegraphics[height=.2\textheight]{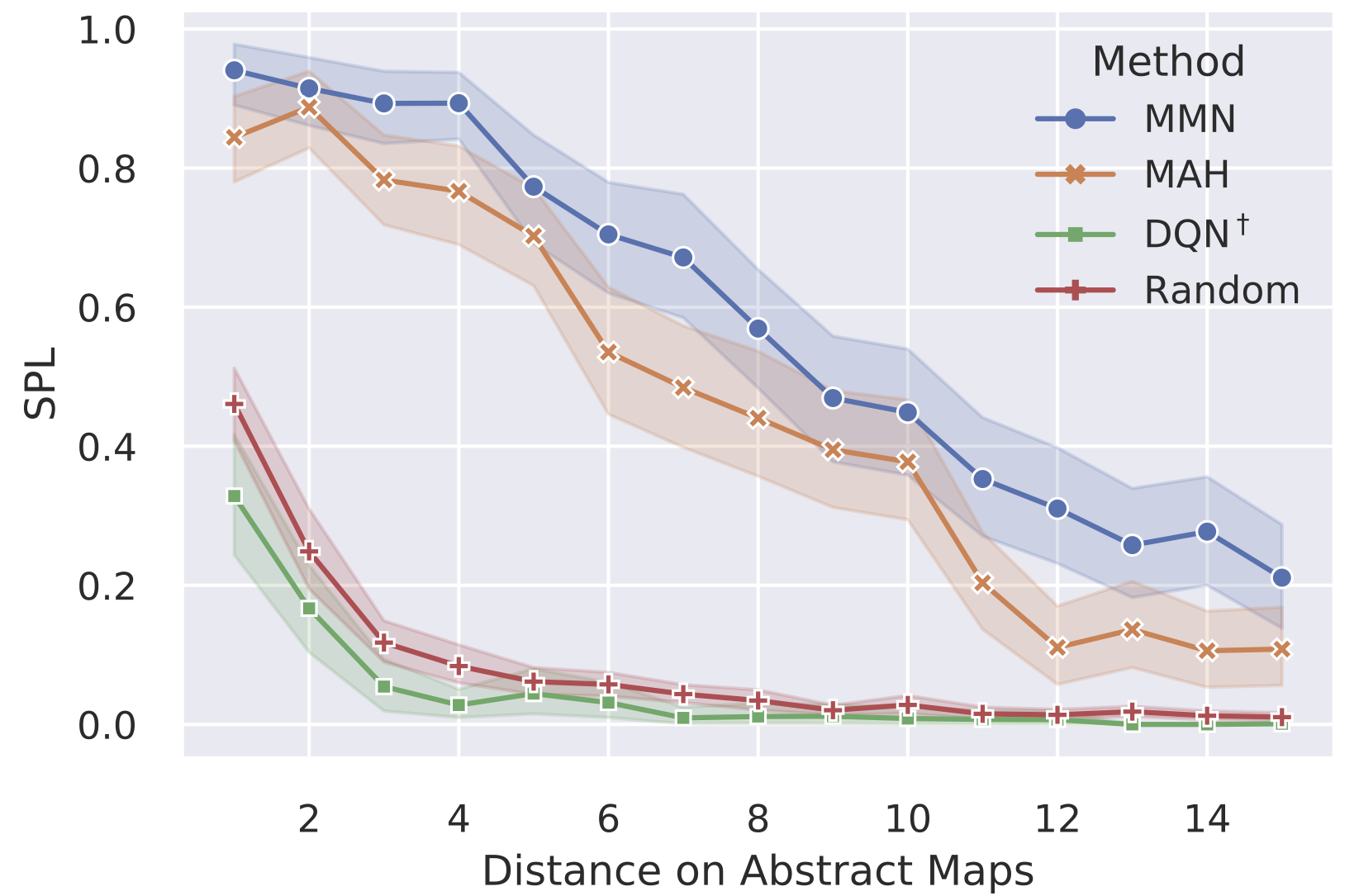}
    \end{subfigure}
    \hfill
    \begin{subfigure}{0.45\linewidth}
	    \includegraphics[height=.2\textheight]{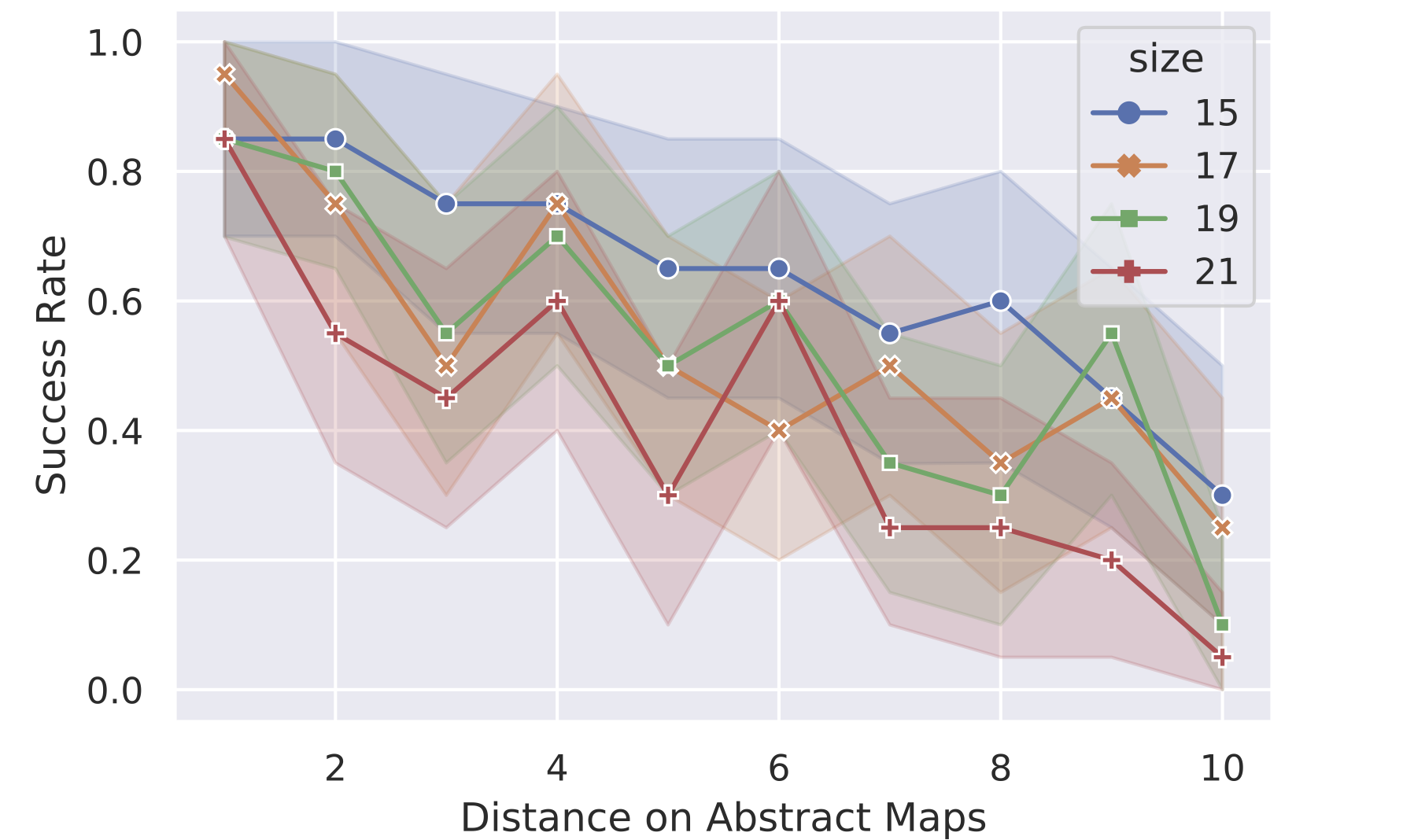}
    \end{subfigure}
    \caption{\small
\textbf{(Left)}
    Zero-shot evaluation performance on $13 \times 13$ maps.
    Local navigation with different distances between start and goal, from 1 to 15.
    \textbf{(Right)} Performance of our method on larger maps.
    }
    \vspace{-10pt}
    \label{fig:combined-zero-shot}
\end{figure}

\subsection{Zero-shot local navigation in novel layouts}

For zero-shot generalization of \textit{locally trained agents}, we \textit{train} all four agents on $20$ of $13 \times 13$ maps with randomly generated local start-goal pairs with distance $[1,5]$ in each episode.
We train the agents until convergence; {\oursmf} typically takes $3 \times$ more training episodes and steps.
We \textit{evaluate} all agents on $20$ \textit{unseen} $13 \times 13$ maps and generate $5$ start-goal pairs for each distance from $1$ to $15$ on each map.
The results are shown in Figure~\ref{fig:combined-zero-shot} left. {\ours} and {\oursmf} generally outperforms the other two baselines. {\ours} has better performance especially over longer distances, both in success rate and successful-trajectory length (not shown), even though it was only trained on distances $\leq 5$.
Since we compare fully trained agents, we found {\ours} performs asymptotically better than {\oursmf}.
{Additionally, as shown in \editiros{Figure~\ref{fig:combined-zero-shot} right}, we also train and evaluate MMN on larger maps from $15\times 15$ to $21\times 21$.
Observed with similar trend to $13 \times 13$ maps, when trained with start-goal distance $\le 5$, the agent will find distant goals and larger maps more difficult.}

\subsection{Hierarchical navigation in novel layouts}
\label{sec:hierarchical}

We also performed a hierarchical navigation experiment, which requires an additional \textit{landmark oracle} to generate sequences of subgoals between long-distance start-goal pairs, and evaluate the performance of hierarchical navigation.
The agent is trained on $13 \times 13$ maps, and evaluate on $20$ unseen $13 \times 13$ maps.
On each map, we use the top-right corner as the global start position and the bottom-left corner as the global goal position, then plan a shortest path in the abstract $2$-D map, and generate a sequence of subgoals with distance $5$ between them; this typically results in $3$ to $6$ intermediate subgoals.
The choice of distance $5$ is motivated by our previous experiment, and because the agent is trained on distances $\leq 5$.
Consecutive subgoal pairs are provided sequentially to the agent as local start-goal pairs to navigate.
The navigation is considered successful only if the agent reaches the global goal by the end.

\begin{figure}[t]
    \centering
    \includegraphics[width=\linewidth]{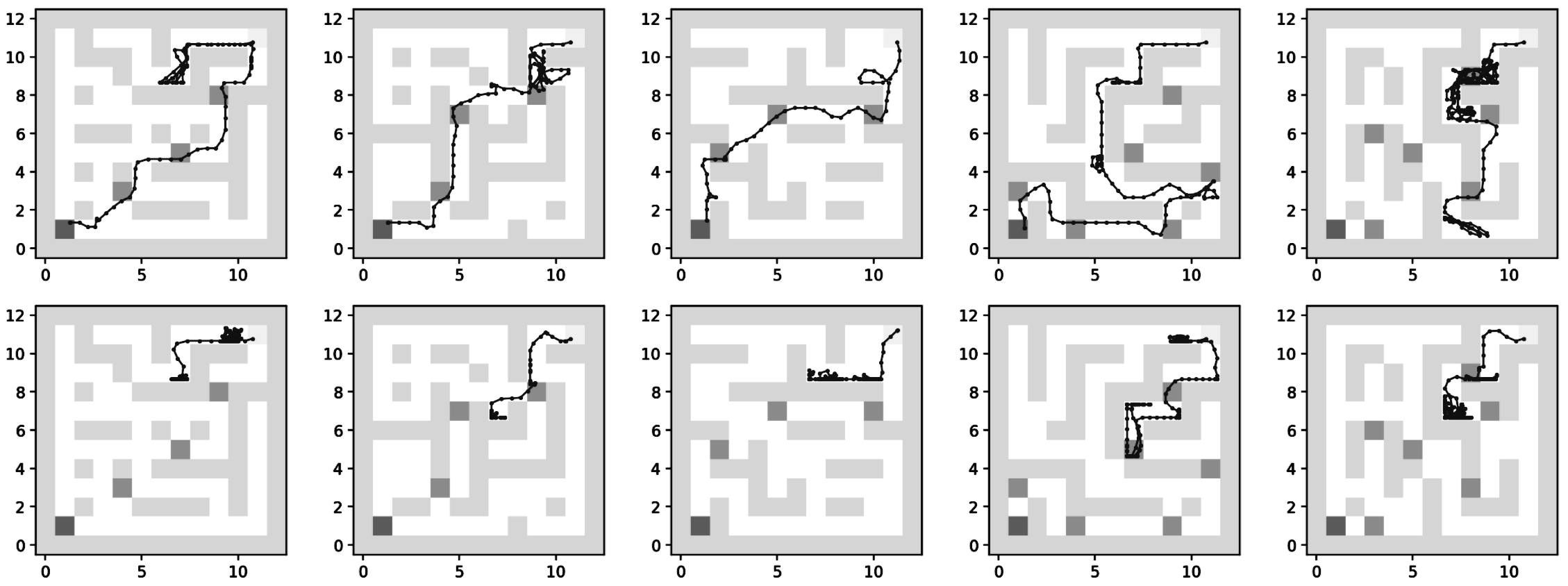}
    \caption{\small
    Trajectories from hierarchical navigation in zero-shot on $13 \times 13$ maps.
    The top row is for {\ours} and bottom row is for {\oursmf}.
    Since there is a fixed scaling factor from maps to environments, we can compute the corresponding location on the abstract map and visualize trajectories,
    although this information is \emph{not} known to the agent.
    The top-right corner is the start, and the bottom-left is the goal. Darker cells indicate provided subgoals from the landmark oracle.
    For the first 4 tasks (columns), {\ours} successfully reached the goals, while {\oursmf} failed. Both methods failed in the last task (right-most column).
    }
    \label{fig:vis-hier}
\end{figure}

We evaluated {\ours} and {\oursmf} on the $20$ evaluation maps.
We provide the next subgoal when the current one is \textit{reached} or until \textit{timeout}.
As shown in Table \ref{tab:hierarchical-nav-res}, our model-based {\ours} outperforms the model-free counterpart by a large margin. {\ours} can reach $16$ out of $20$ global goals, which include all $9$ successful cases of {\oursmf}.
We visualize five trajectories of zero-shot hierarchical navigation in \editiros{Figure~\ref{fig:vis-hier}}. The model-based {\ours} is more robust to the intermediate failed subgoals by navigating to the new subgoal directly, where the model-free {\oursmf} gets stuck frequently.

\begin{table}[h]
 \centering
     \caption{\small
     Hierarchical navigation performance for various distances between the \textit{landmarks}, measured by SPL \editiros{and success rate (SR only shown for distance 5)}. 
    Landmarks are provided subgoals between fixed start-goal pairs on 20 maps. SPL performance is not monotonic because it reflects (lack of) optimality.
     }
    \label{tab:hierarchical-nav-res} 
    \begin{tabular}{c|rrrrr|r}
        Landmark Distance & 1 & 2 & 3 & 4 & 5 & \editiros{5 (SR)} \\
      	\hline
      	\textbf{\ours} & \textbf{0.61} & \textbf{0.59} & \textbf{0.68} & \textbf{0.45} & \textbf{0.63} & \textbf{0.80} \\
      	\textbf{\oursmf} & 0.24 & 0.42 & 0.45 & 0.41 & 0.28 & 0.45 \\
      	\textbf{DQN}$^\dagger$ & 0.00 & 0.00 & 0.00 & 0.00 & 0.00 & 0.00 \\
      	\textbf{Random} & 0.00 & 0.00 & 0.00 & 0.00 & 0.00 & 0.00 \\
    \end{tabular}
\end{table}

In Appendix~\ref{sec.appendix-results}, we provide further experiments studying the robustness of our method to various perturbations, including situations where the abstract map contains inaccuracies and where the agent is only provided a noisy version of its location. In general, our learning-based agent is robust to these changes, though performance gradually degrades as the magnitude of perturbation increases.

\section{Conclusion}
\label{sec:conclusion}

\editiros{
In this work, we have presented an end-to-end model-based approach, {\ours}, for enabling agents to navigate in environments with novel layouts. By using provided abstract $2$-D maps and start/goal information, MMN does not require further training or exploration (zero-shot).
Compared to the map-conditioned model-free counterpart {\oursmf},}
both approaches performed well in zero-shot navigation for short distances;
for longer distances (with access to a landmark oracle), our model-based approach {\ours} performed significantly better.
In future work, we will explore learned subgoal generators, handle visual observation input, and perform navigation in rich visual environments.




\bibliography{main}
\bibliographystyle{rlc}


\newpage
\appendix

\section{Further details on our approach}
\label{sec.appendix-approach}

\subsection{$n$-step goal relabelling: Denser reward}

Jointly training a planner with learned model can suffer from lack of reward signal, especially when the model training entirely relies on reward from \textit{multiple tasks}, which is common in model-based agents based on \textit{value gradients} \citep{schrittwieser2019mastering,oh2017value}.
Motivated by this, we introduce a straightforward strategy to enhance the reward signal by implicitly defining a learning curriculum, named \textit{$n$-step hindsight goal relabelling}.
This generalizes the single-step version of \textit{Hindsight Experience Replay} (HER) \citep{andrychowicz2017hindsight} to \textit{$n$-step return relabeling}. 

\editiros{
\textit{Motivation.}
}
As shown in Figure~\ref{fig:inference-training} (right), we sample a trajectory of experience $\left(c_\mathcal{T}, \left\{ s_t, a_t, r_t, s_{t+1} \right\}_t \right)$ on a specific map and goal $c_\mathcal{T}=(m_\mathcal{T}, g_\mathcal{T})$ from the replay buffer.
Observe that, if the agent does not reach the goal area $\mathcal{S}_\mathcal{G}$ (a $100 \times 100$ cell in the agent's $3$-D environment, denoted by a $2$-D position $g_\mathcal{T}$ on the abstract 2-D map), it will only receive reward $r_t=-1$ during the entire episode until timeout.
In large maps, this hinders the agent to learn effectively from the current map $m_\mathcal{T}$. Even if the agent partially understands a map, it would rarely experiences a specific goal area on the map again.\footnote{In our \textit{extremely} low data regime, the agent only has one start-goal pair on a small set of map. While on \textit{low} data regime, the agent can train on randomly sampled pairs on the maps. See the Setup for more details.}
This is more frequent on larger maps in which possible navigable space is larger.

\editiros{
\textit{Relabelling $n$-step returns.}
}
Motivated by single-step HER, we relabel failed goals to randomly sampled \textit{future} states (visited area) from the trajectory, and associating states with the relabelled $n$-step return.
Concretely, the \textit{task-conditioned} bootstrapped $n$-step return is
\begin{equation}
G_{t}^\mathcal{T} \doteq r_{t+1}+\gamma r_{t+2}+\cdots+\gamma^{n} v_{n}^\mathcal{T}, \quad \left[ v_{n}^\mathcal{T}, \pi_n^\mathcal{T} \right] = g_\theta(s_t, c_\mathcal{T})
\end{equation}
where $v_{n}^\mathcal{T}$ is the state-value function \textit{bootstrapping} $n$ steps into the future from the search value and \textit{conditional} on task context $c_\mathcal{T}$.
This task-conditioned value function is \textit{asymmetric} since $\mathbb{R}^{12} = \mathcal{S} \ne \mathcal{S}_g = \mathbb{R}^2$.

\editiros{
\textit{Steps.}
}
To relabel the task-conditioned bootstrapped $n$-step return, there are three steps, \editiros{demonstrated by the blue lines from ``$N$-step Relabel'' box.}
(1) \textit{Goal \editiros{(red boxes)}.} Randomly select a reached state $s_t \in \mathbb{R}^{12}$ from the trajectories, then take the $2$-D position $(x,y) \in \mathbb{R}^2$ in agent world and convert it to a $2$-D goal support grid $g_{\mathcal{T}_S}$.
Then, relabel the \textit{goal} in task context $c_{\mathcal{T}_S}=(m_\mathcal{T}, g_{\mathcal{T}_S})$, keeping the abstracted map and start position unchanged.
(2) \textit{Reward \editiros{(orange boxes)}.} Recompute the rewards along the n-step segment. In episodic case, we need to terminate the episode if the agent can reach the relabelled goal area $g_{\mathcal{T}_S}$, by marking "done" at the certain timestep or assigning zero discount after that step $\gamma_t = 0$ to mask the remaining segment.
(3) \textit{Value \editiros{(purple circles)}.} Finally, we need to recompute the bootstrapping task-conditioned value $v_{n}^{\mathcal{T}_S}, \pi_n^{\mathcal{T}_S} = g_\theta(s_t, c_{\mathcal{T}_S})$.

Empirically, this strategy significantly increases the efficiency of our multi-task training by providing smoothing gradients when sampling a \textit{mini-batch} of $n$-step targets from successful or failed tasks.
It can also be applied to other multi-task agents based on $n$-step return.

\subsection{Joint optimization: Multi-task value learning}

Our training target has two components. The first component is based on \editiros{the value gradient objective in MuZero \citep{schrittwieser2019mastering,oh2017value}, using relabelled experiences from proposed $n$-step HER.
It is denoted by $\mathcal{L}_{\text {task }}^k$ for step $k = 1, \ldots, K$.
}
\editiros{
However, this loss is only suitable for single-task RL.
}

\editiros{
Thus, we propose an auxiliary \editiros{model prediction} loss, denoted by $\mathcal{L}_{\text {model}}^k$ in Figure~\ref{fig:inference-training} (right).
The motivation is to regularize that the hypermodel $f_\phi(s,a, h_\psi(c_\mathcal{T}))$ should predict trajectory based on the information of given abstract map and goal $c_\mathcal{T}$.
}
The objective corresponds to maximizing the mutual information between task context $c_\mathcal{T}$ and \textit{predicted} trajectories $\hat{\tau}_\mathcal{T}$ from the hypermodel on sampled tasks $\mathcal{T} \sim \rho (\mathcal{T})$:
\begin{equation}
    \max\limits_{h_\psi} \mathbb{E}_{\mathcal{T} \sim \rho (\mathcal{T})} \left[ I(c_\mathcal{T}; \hat{\tau}_\mathcal{T}) \right],
\end{equation}
where $h_\psi(c_\mathcal{T})=\phi$ is the meta network predicting the weight of transition network $f_\phi$. 
Observe that: $I(\tau; c) = H(\tau) - H(\tau | c) \ge H(\tau) + \mathbb{E}_{\tau, c} \left[ \log q(\tau | c) \right]$, we can equivalently optimize the RHS $\max_h \mathbb{E}_\mathcal{T} \left[ \log q(\tau | c) \right] \iff \max_h \mathbb{E}_{(s,a,s') } \left[ \log q(s' | s, a ; h(c)) \right]$ (subscripts omitted).
\editiros{
This objective is equivalent to minimizing the loss between predicted states and true states from environment, for all transition tuples across all tasks.
}
\editiros{The final loss is given by the sum over multiple steps:
\begin{equation}
\mathcal{L}(\psi, \phi, \theta) = \sum\limits_{k=1}^{n} \mathcal{L}_{\text {task}}^k + \mathcal{L}_{\text {model}}^k,
\end{equation}
where $k = 1, ..., K$, and $K$ is the length of training segment.
}

\section{Further experiments and results}
\label{sec.appendix-results}

\subsection{Robustness to map and localization errors}
\label{sec:perturbation}

To further study the robustness of our method and the importance of each component, we considered breaking three components in closed-loop map-based navigation: Map -- (1) $\rightarrow$ Path -- (2) $\rightarrow$ Environment -- (3) $\rightarrow$ Map (repeat). In general, our learning-based agent is robust to these changes.
To illustrate the difficulty of the problem, we considered a hard-coded strategy (hand-crafted deterministic planner) based on perfect information of the environment (e.g., can plan on map) for comparison correspondingly: (1) known perfect maps and intermediate landmarks, (2) scaling factor (unavailable to MMN), and (3) world position on map.
Since we assume that it has perfect localization and landmarks, the key step is to reach a landmark given current location, which consists of several procedures: (a) change the orientation to the target landmark, (b) move forward along the direction, and (c) stop at the target cell as soon as possible.

\begin{figure}[h]
    \centering
    \hfill
    \begin{subfigure}{0.45\linewidth}
       	\includegraphics[width=\linewidth]{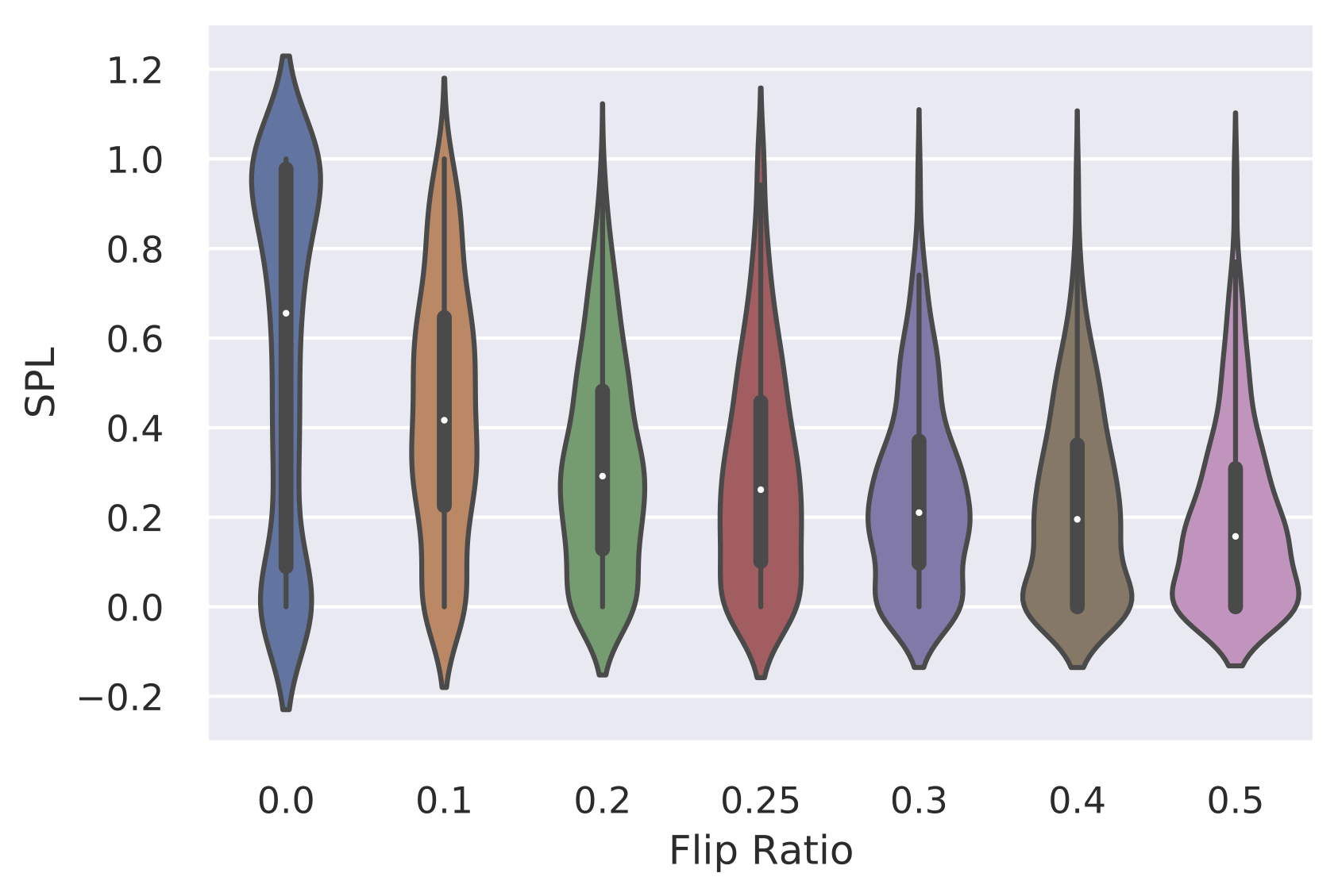}
    \end{subfigure}
    \hfill
    \begin{subfigure}{0.45\linewidth}
        \includegraphics[width=\linewidth]{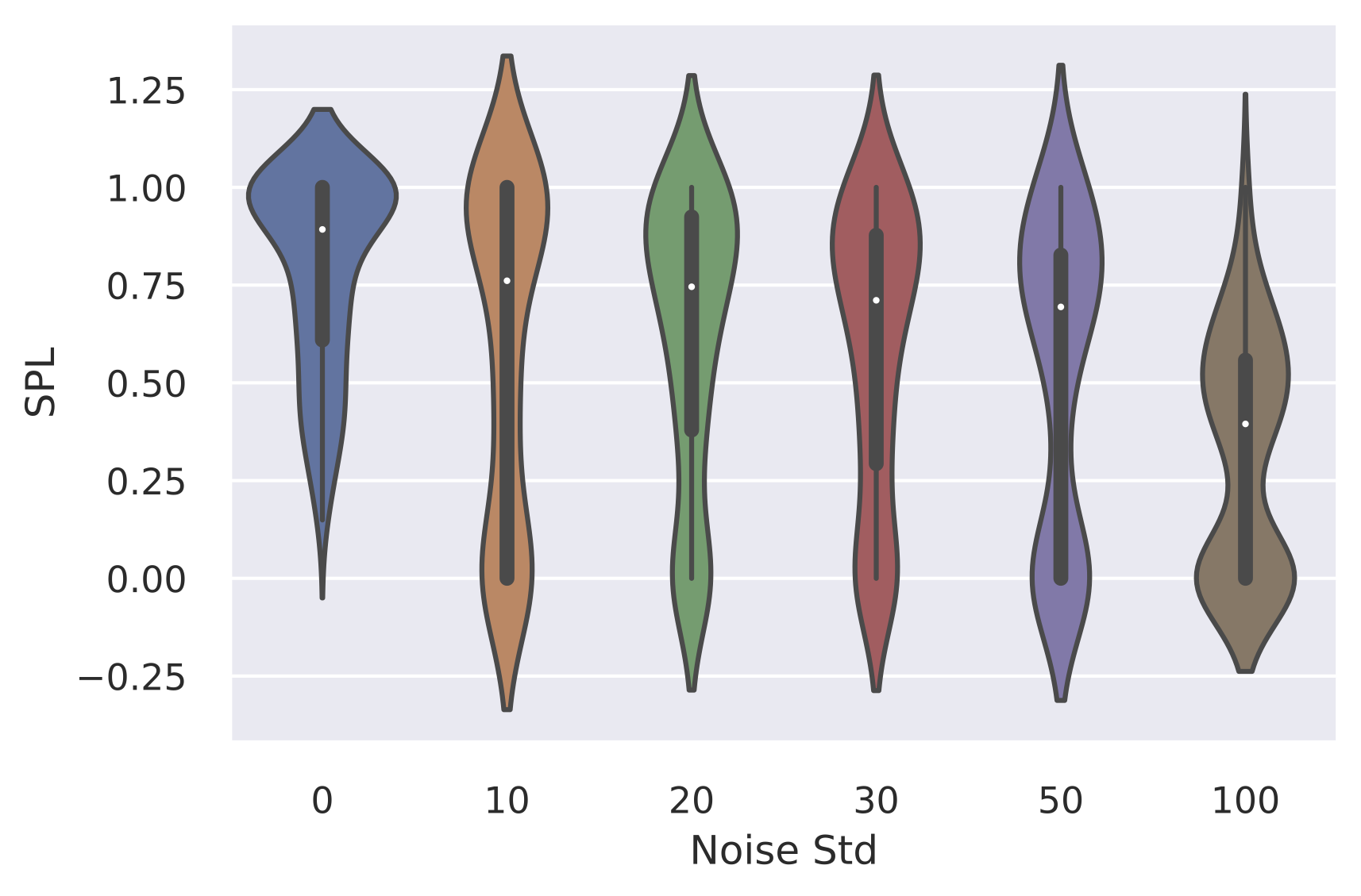}
    \end{subfigure}
    \hfill
    \caption{
 	Violin plots show the SPL of MMN with different map flip ratio \textbf{(left)} and localization noise level \textbf{(right)}. The two figures clearly show the negative impact of imperfect information, which also justify the importance of the guidance.
    }
    \label{fig:further-study-main-paper}
\end{figure}

\paragraph{\editmark{Perturbing planning}}
We try to break the implicit assumption of requiring perfect abstract map information. We adopt the hierarchical setting, but generating subgoals on perturbed maps, where some proportion of the map's occupancy information is flipped.
In Figure~\ref{fig:further-study-main-paper} (left), as the perturbation level increases, MMN's performance gradually decreases, but it still navigates successfully with significant noise levels.

\paragraph{\editmark{Perturbing action mapping}}
We break the implicit requirement of known scaling between map and environment. We provide the agent with randomly transformed maps with random perspective transformation, where the ratio (in both x and y directions) is different.
\editmark{As shown in Table~\ref{tab:action-mapping-res}}, perturbed MMN's performance decreases gracefully compared to unperturbed one, which shows that our agent rely little on this knowledge or any perfect relation.

\begin{table}[h]
 \centering
     \caption{
     Success rate for perturbing action mapping, comparing with unperturbed MMN for reference.
     }
    \label{tab:action-mapping-res}
    \begin{tabular}{l|rrrrr}
    \hline
        Goal Distance & 2 & 4 & 6 & 8 & 10 \\
      	\hline
      	\textbf{MMN} (Perturbed) & 0.80 & 0.85 & 0.71 & 0.40 & 0.36 \\
		\textbf{MMN} (Default) & 0.91 & 0.90 & 0.71 & 0.58 & 0.43 \\
        \hline
    \end{tabular}
\end{table}

\begin{figure}[h]
    \centering
    \includegraphics[width=0.2\textwidth]{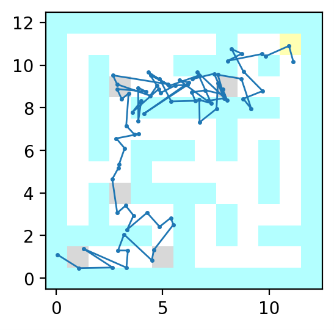}
    \includegraphics[width=0.2\textwidth]{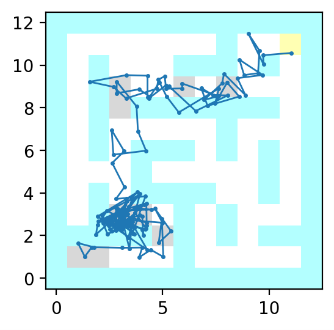}
    \qquad
    \includegraphics[width=0.2\linewidth]{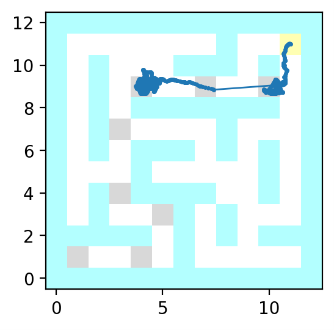}
    \includegraphics[width=0.2\linewidth]{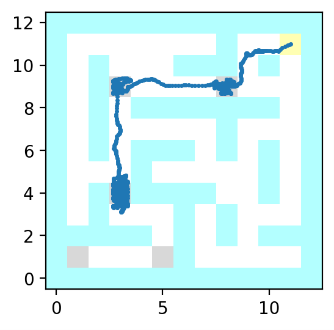}
    \caption{
	(\textbf{left pair}) MMN visualized with \textit{perturbed} locations; even though the provided state is noisy, MMN successfully reaches the goal.
	(\textbf{right pair}) Deterministic planner is unable to reach the goal when the provided state is noisy. (We only show the \textit{unperturbed} locations in this case for clarity in visualization.)
	MMN still reaches the goal with 50 units of noise (0.5 cell), while the deterministic planner gets stuck at some subgoals or runs out of budget.
    }
    \label{fig:localization-perturbation-compare}
\end{figure}

\paragraph{\editmark{Perturbing location}}
We break the identifiability of agent position (a part of its joint state) by applying random noise to given position. We aim to show that our agent does not rely on the position to understand the map, since providing position in the agent world has no relation with localizing on abstract maps and our learning-based method can adapt to the noise.
In Figure~\ref{fig:further-study-main-paper} (right), even though MMN is trained without noise, it tolerates some amount of noise and maintains relatively high SPL even at 50 units of noise (corresponding to 0.5 cell width).
{In Figure~\ref{fig:localization-perturbation-compare}, we visualize the trajectories of MMN and the deterministic planner to qualitatively demonstrate MMN's robustness to noise.}

\end{document}